\documentclass{mva_style}
\usepackage{graphicx}
\usepackage{subcaption}
\usepackage{float}
\usepackage[cc]{titlepic}
\usepackage{cite}
\usepackage{arydshln}
\usepackage{hyperref}

\begin{document}
\title{\textbf{Self-Supervised Pre-Training Boosts \\ Semantic Scene Segmentation on LiDAR data}}

\author{
  Mariona Carós$^{1}$\thanks{This research was partially funded by 2021 DI 41 and 2021 SGR 01104 (Generalitat de Catalunya)}, Ariadna Just$^{2}$, Santi Seguí$^{1}$, Jordi Vitrià$^{1}$\\
  $^{1}$Dept. de Matemàtiques i Informàtica, Universitat de Barcelona, Barcelona\\
  $^{2}$Cartographic and Geological Institute of Catalonia, Barcelona
}

\titlepic{\includegraphics[width=\textwidth]{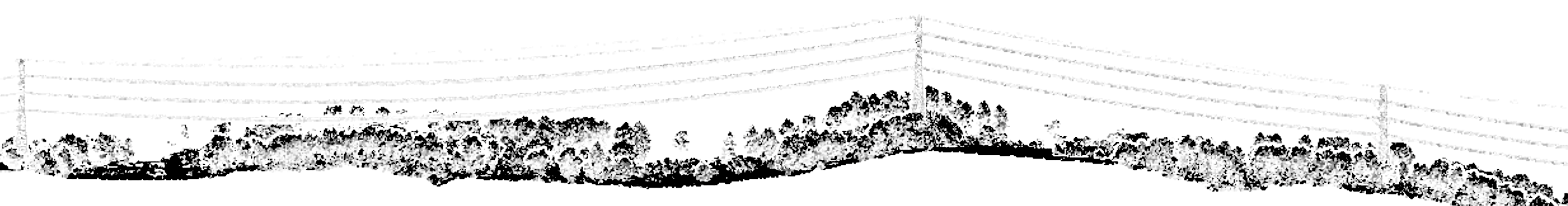}}

\date{\vspace{-5ex}} 

\maketitle

\section*{\centering Abstract}
\textit{ 
Airborne LiDAR systems have the capability to capture the Earth's surface by generating extensive point cloud data comprised of points mainly defined by 3D coordinates. However, labeling such points for supervised learning tasks is time-consuming. As a result, there is a need to investigate techniques that can learn from unlabeled data to significantly reduce the number of annotated samples. In this work, we propose to train a self-supervised encoder with Barlow Twins and use it as a pre-trained network in the task of semantic scene segmentation. The experimental results demonstrate that our unsupervised pre-training boosts performance once fine-tuned on the supervised task, especially for under-represented categories.}

\section{Introduction}

Airborne LiDAR (Light Detection And Ranging) is a remote sensing technology that employs near-infrared light to produce highly accurate three-dimensional (3D) representations of the Earth's surface, as exemplified in the header image. The number of points in a scene acquired by a LiDAR sensor is usually immense, typically comprising millions of points. While labeling can be automated for some elements, such as ground and planar surfaces like buildings, some objects require manual annotation due to their varied shapes and relatively low representation, often comprising less than 1\% of the points in the data. Consequently, labeling point cloud data is a task that requires a significant amount of time and effort which results in a lack of large annotated 3D datasets. 

 
In this work, we aim to use self-supervised learning (SSL) on unlabeled point clouds, inspired by the success of self-supervised methods in natural language processing\cite{dai2015semi, devlin2018bert, radford2019language} and computer vision \cite{larsson2017colorization, caron2018deep, grill2020bootstrap, He_2020_CVPR}, to obtain meaningful representations for a semantic scene segmentation task. 
Point cloud semantic segmentation, also referred to as point classification, involves the prediction of a categorical label for every point within a given point cloud. This task is especially challenging due to the scattered and irregular nature of aerial LiDAR data, which comprises an extensive number of points. Several architectures~\cite{qi2017pointnet, hu2022point, chen2017multi, landrieu2018large} have been implemented to process point cloud data, including point-based networks, graph-based networks, voxel-based networks, and multi-view networks. Since LiDAR sensors acquire data in the form of 3D points, our focus is on exploring the efficacy of point-based networks for this task. The pioneering work to directly process point cloud data was PointNet~\cite{qi2017pointnet}. Qi et al.~\cite{qi2017pointnet++} extended the capabilities of PointNet by incorporating local geometric information through a hierarchical neural network, which resulted in PointNet++. Inspired by the mentioned networks, recent studies~\cite{xie2020linking, Zhao_2021_ICCV, caros2022object} focus on redefining sampling and augmenting features using knowledge from other fields to improve its performance. 

Self-supervised pre-training has emerged as a promising technique for supervised tasks like image segmentation and classification in situations where access to annotations is limited. A successful strategy involves learning embeddings that remain invariant to input data distortions by maximizing the similarity of representations subject to different conditions\cite{chen2020simple, chen2021exploring}.
In this context, methods differ in the similarity function used, whether the encoders for input samples are the same or different, and the type of transformations utilized. Notable examples are contrastive methods, such as SimCLR \cite{chen2020simple}, clustering approaches \cite{caron2018deep, caron2020unsupervised}, and Siamese networks \cite{chen2021exploring}. Our approach is based on Barlow Twins~\cite{zbontar2021barlow} which minimizes redundancy via cross-correlation between outputs of two identical networks fed with distorted versions of a sample. While solving this task, representations that capture semantic properties of the point cloud are learned.
Barlow Twins does not fall under the categories of either contrastive learning or clustering methods. Its design provides several benefits, such as not requiring large batches \cite{chen2020simple}, asymmetric mechanisms \cite{grill2020bootstrap}, or stop-gradients \cite{chen2021exploring}.

Recent advances in SSL for 2D data have motivated research in applying similar techniques to 3D processing. For instance, PointContrast  \cite{xie2020pointcontrast} leverages multi-view depth scans with point correspondences for high-level scene understanding tasks. However, this method is limited to static scenes that have been registered with two views. Other works \cite{achlioptas2018learning, sauder2019self, wang2019deep, hassani2019unsupervised} directly feed point cloud data into the network for SSL, although most of them focus on single 3D object representation for reconstruction, classification, or part segmentation. Few studies include scene representations \cite{zhang2021self, xie2020pointcontrast}, and these mainly focus on indoor and driving scenes provided by terrestrial laser scanners.

In order to improve performance across real-world tasks through SSL, exploring strategies on single objects may present limited potential. Hence, we propose pre-training the network on complex scenes obtained by LiDAR to better match the target distributions. To the best of our knowledge, this study is novel in utilizing a self-supervised method such as Barlow Twins to pre-train a neural network for the task of scene segmentation using airborne LiDAR data. The code of this study is publicly available at \href{https://github.com/marionacaros/barlow-twins-for-sem-seg}{github.com/marionacaros/barlow-twins-for-sem-seg}. Our contributions can be summarized as follows: 

\begin{itemize}

\item We propose a methodology for pre-training a 3D semantic scene segmentation model by using SSL.

\item We show that SSL can be used with sparse outdoor LiDAR data, even if the dataset is highly imbalanced.

\item We experiment with PointNet and PointNet++ and show a significant performance improvement in semantic scene segmentation over under-represented categories within different datasets.

\end{itemize}

\section{Method}

\begin{figure}[t]
\centering
\includegraphics[width=\columnwidth]{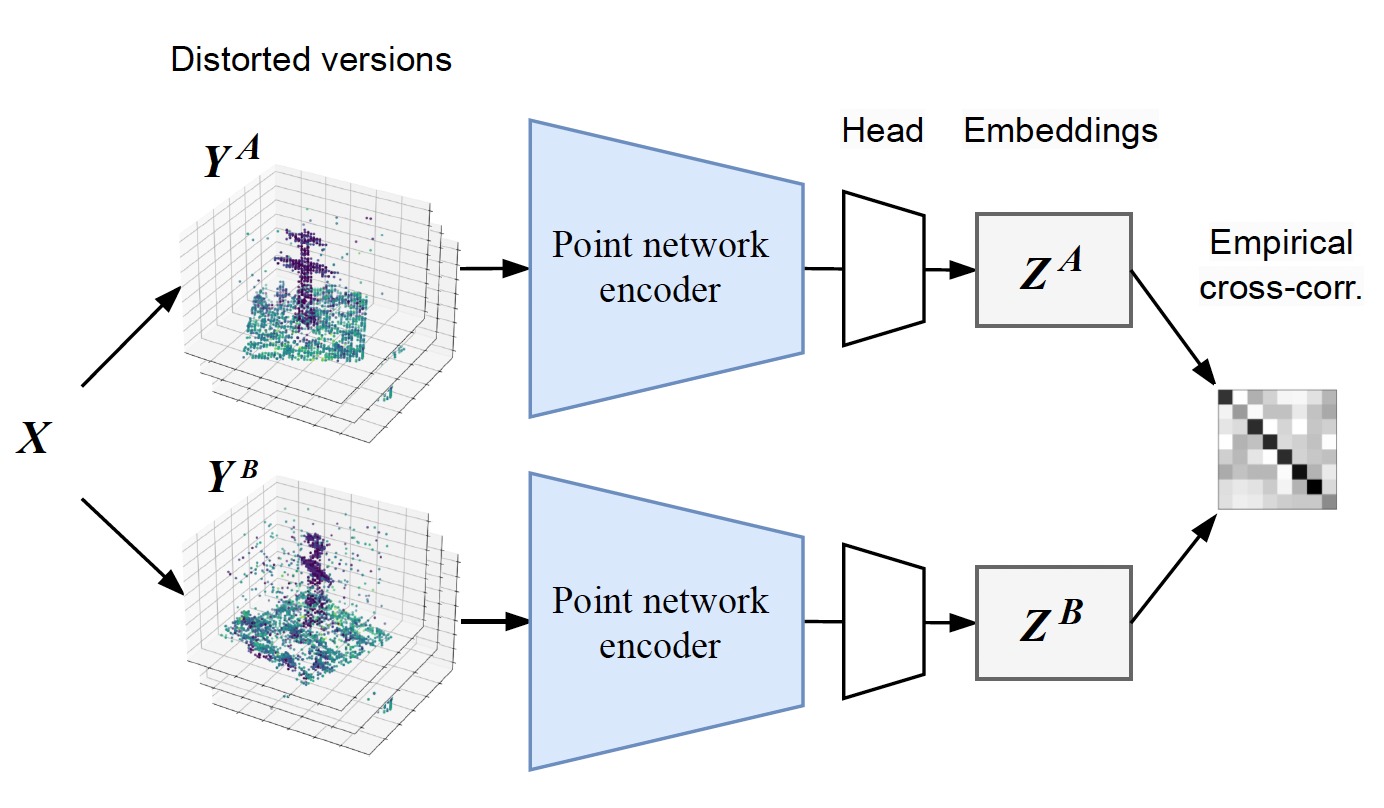}
  \caption{Barlow Twins method consists of: (i) Producing two distorted views for all point clouds of a batch. (ii) Feeding batches to two identical deep encoder networks producing embeddings. (iii) By using the objective function, the cross-correlation between features is optimized to be the identity. 
  }
\label{fig:barlow}
\end{figure}

We introduce our approach for semantic scene segmentation given a small portion of labeled data. 
The methodology consists in training a self-supervised point-based network on point clouds and using it as initialization for the supervised task.
The SSL method is illustrated in Fig~\ref{fig:barlow}. Based on Barlow Twins architecture \cite{zbontar2021barlow}, it applies redundancy-reduction by using a joint embedding of distorted point cloud views, which learns powerful representations from LiDAR data.


\subsection{Self-Supervised Training}

Given a dataset of partially labeled point clouds, we  wish to effectively utilize all the available data to train a neural network that can accurately perform semantic scene segmentation. We begin by training a supervised network for this task on the labeled data. Specifically, we use the point-based architectures PointNet \cite{qi2017pointnet} and PointNet++\cite{qi2017pointnet++} due to their simplicity and efficiency.

Next, we split the dataset into $N$ batches ${D=\{X_i\}_{i=1}^N}$. To each batch $X_i$, we apply a data augmentation to obtain two distorted versions. Let $t_A, t_B \in \mathcal{T}$, be randomly sampled augmentations from a set of transformations. Consequently, we can define the two batches of distorted point clouds as ${Y^A=t_A(X)}$ and ${Y^B=t_B(X)}$. $Y^A$ and $Y^B$ are fed to the encoder network, which is used as initialization for Barlow Twins method. Then, a projection head is used after the encoder to map representations to the space where the objective function is applied. Finally, the network output are batches of embeddings $Z^A$ and $Z^B$ which are used by the Barlow Twins objective function to measure its cross-correlation matrix. The objective function is defined as follows:

\begin{equation}
Loss_{BT}=\sum_{i}(1-C_{ii})^2 + \lambda \sum_{i}\sum_{j\neq i}C_{ij}^2
\end{equation}
where $\lambda$ is a positive constant weighing the components of the loss, and where $C$ is the cross-correlation matrix computed between the embeddings $Z^A$ and $Z^B$.

The cost function is composed of two terms. The first term aims to make the embedding invariant to the applied distortion by trying to equate the diagonal elements of the cross-correlation matrix to 1. While the second term tries to equate the off-diagonal elements of the cross-correlation matrix to 0, reducing the redundancy between the output units. 

Once the encoder is pre-trained on unlabeled training data, we train it for semantic scene segmentation.

\subsection{Point Cloud Distortions}

\begin{figure}[h]
    \centering
    \begin{subfigure}[h]{0.45\columnwidth}
      \includegraphics[width=\linewidth]{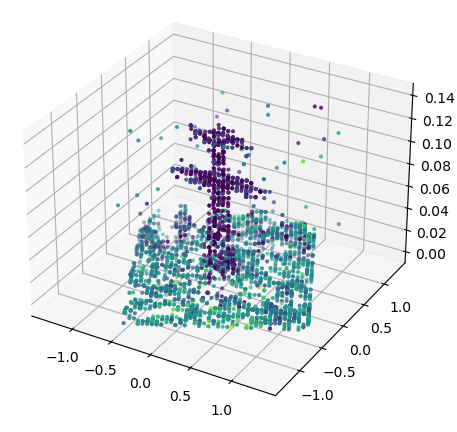}
        \caption{Version 1}
        \label{fig:dist1}
    \end{subfigure}
    \begin{subfigure}[h]{0.45\columnwidth}
   \includegraphics[width=\linewidth]{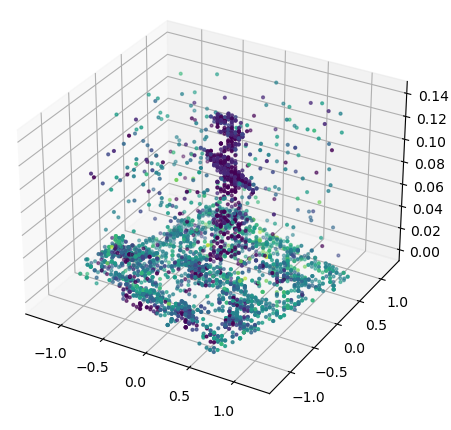}
        \caption{Version 2}
        \label{fig:dist2}
    \end{subfigure}
    \caption{Augmented point clouds.}
    \label{fig:dist_plots}
\end{figure}

Each input point cloud is transformed twice to produce two distorted views, an example of them is shown in Fig~\ref{fig:dist_plots}.
We first apply typical data augmentation strategies for point clouds which are: random down-sampling, up-sampling by point duplication, and random rotation in $xy$ axis. Then, we add a new transformation which involves moving a percentage of points by modifying their coordinates with random values, while preserving other attributes. The goal of this transformation is to introduce noise into the point cloud while maintaining its dimension and preserving the shape of objects within it. To achieve this, points are randomly chosen with probabilities ranging from 2\% to 5\% in the first version, and a fixed probability of 10\% in the second version. We consider that higher percentages may excessively alter the shape of certain objects.

\subsection{Architecture}

\textbf{Self-supervised learning} \quad
We use Barlow Twins which is comprised of two networks: an encoder and a projector. The encoder is implemented as a point-based network, wherein the final layers responsible for classification are omitted. The projector is constructed using two linear layers with 512 and 128 units. The first linear layer is followed by a batch normalization layer and a Rectified Linear Unit (ReLU) activation function. The output of the encoder is used for transfer tasks whose dimension depends on the size of the encoder network. 
The output of the projector is an embedding of 128 which is fed to the loss function for optimization of Barlow Twins. 
 
\textbf{3D Semantic segmentation} \quad
We experiment with PointNet and PointNet\texttt{++} as our supervised methods for scene segmentation. 
PointNet uses multi-layer perceptrons (MLPs) to learn local features corresponding to each point and a symmetric function to obtain a representation of the point cloud. PointNet encoder applies input and feature transformations by using a transformation net~\cite{jaderberg2015spatial}, and then aggregates point features by max pooling producing a global point cloud embedding. The segmentation network concatenates global and local features to produce per-point scores. The details of the architecture can be found in the original work \cite{qi2017pointnet}. PointNet\texttt{++}\cite{qi2017pointnet++} is a hierarchical neural network that applies PointNet recursively on a nested partitioning of the input point set, which enables capturing local structures in point clouds. Both networks take 4096 points as input. 


\section{Experimental Setup}

\subsection{Dataset}

We evaluate the performance of our method on two airborne LiDAR datasets: The DALES benchmark dataset \cite{varney2020dales}, which is a publicly available dataset collected from an aerial laser scanner (ALS) in the city of Dayton; and a private dataset named LiDAR-CAT3, which we describe in greater detail in this section.


The LiDAR-CAT3 dataset was collected by a Terrain Mapper 2 system, which combines a LiDAR sensor with two nadir cameras in RGB and NIR (Near InfraRed). The obtained average point density of first returns is 10 ppm. We use 8 of the attributes provided by the system: coordinates $(x, y, z)$, intensity, three color channels (R, G, B), and NIR channel. Additionally, we compute the NDVI (Normalized Difference Vegetation Index) \cite{pettorelli2013normalized}. The study area is composed of hilly and densely forested zones in Spain, where the orthometric heights were normalized to height above ground to account for changes in altitude due to mountainous terrain. Then, ground points ($z=0$) and outliers ($z>100$ m) were filtered. Ground filtering is a common practice in ALS data processing~\cite{mandlburger2017improved}, as the number of ground points in a scene often surpasses that of object points. Finally, we divided the area into point clouds of 40 m $\times$ 40 m $\times$ 50 m to be used as input samples to our networks. The classified categories with their relative percentage within labeled data are: high vegetation (74.88\%), low vegetation (23.06\%), roofs (2.01\%), pylon (0.02\%), wires (0.02\%), and other buildings (0.01\%).

\subsection{Pre-processing}

In consideration of the highly dense vegetation areas within LiDAR-CAT3 dataset, we employed SemDeDup\cite{abbas2023semdedup} for reducing redundancy prior to training our SSL network. This approach involves removing redundant samples from the dataset by using pre-trained model embeddings to identify data samples that are semantically similar. Specifically, we utilized the embeddings generated by PointNet, previously trained on labeled data. The similarity threshold was set to 0.996, which reduced our unlabeled training dataset by 60\%, increasing the proportion of under-represented categories within the dataset. We found this step essential for effectively learning representations of minority classes.

\subsection{Implementation Details}

We train\footnote{GPU: NVIDIA RTX A6000 - 48 GB} our Barlow Twins network for 300 epochs with a batch size of 150. We use a learning rate of $10^{-4}$ and a linear warm-up schedule period of 10 epochs. For optimization of the supervised point-based networks, we set the initial learning rate to $10^{-3}$, with a decay rate of 0.5 every 50 epochs. The optimizer used is Adam. Supervised networks are trained for 100 epochs with an early stopping on the validation loss. As for the batch size, it is set to 32. The loss is the weighted cross-entropy with double weight on low-represented categories ($<$1\% data points).  


\section{Experiments and Results}
\label{sec:experiments}

\begin{figure}[t]
    \centering
    \begin{subfigure}[h]{0.4\columnwidth}
      \includegraphics[width=\linewidth]{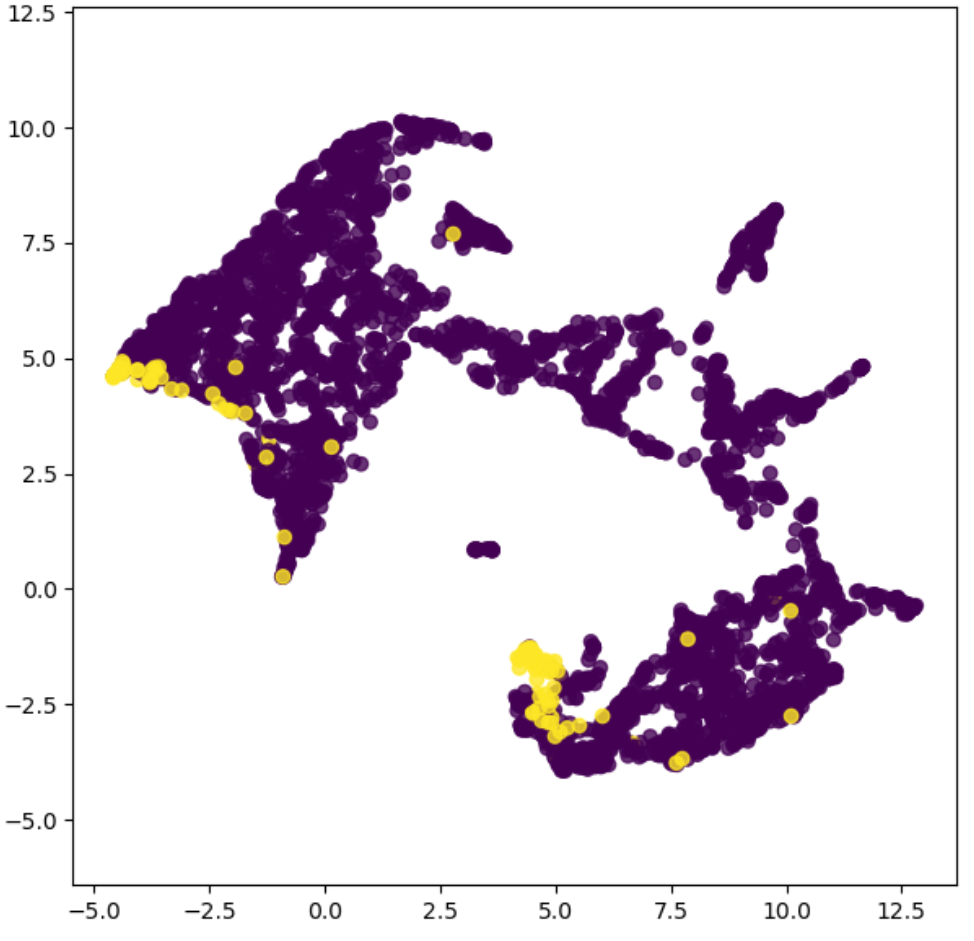}
        \caption{UMAP}
        \label{fig:umap}
    \end{subfigure}
    \begin{subfigure}[h]{0.4\columnwidth}
   \includegraphics[width=\linewidth]{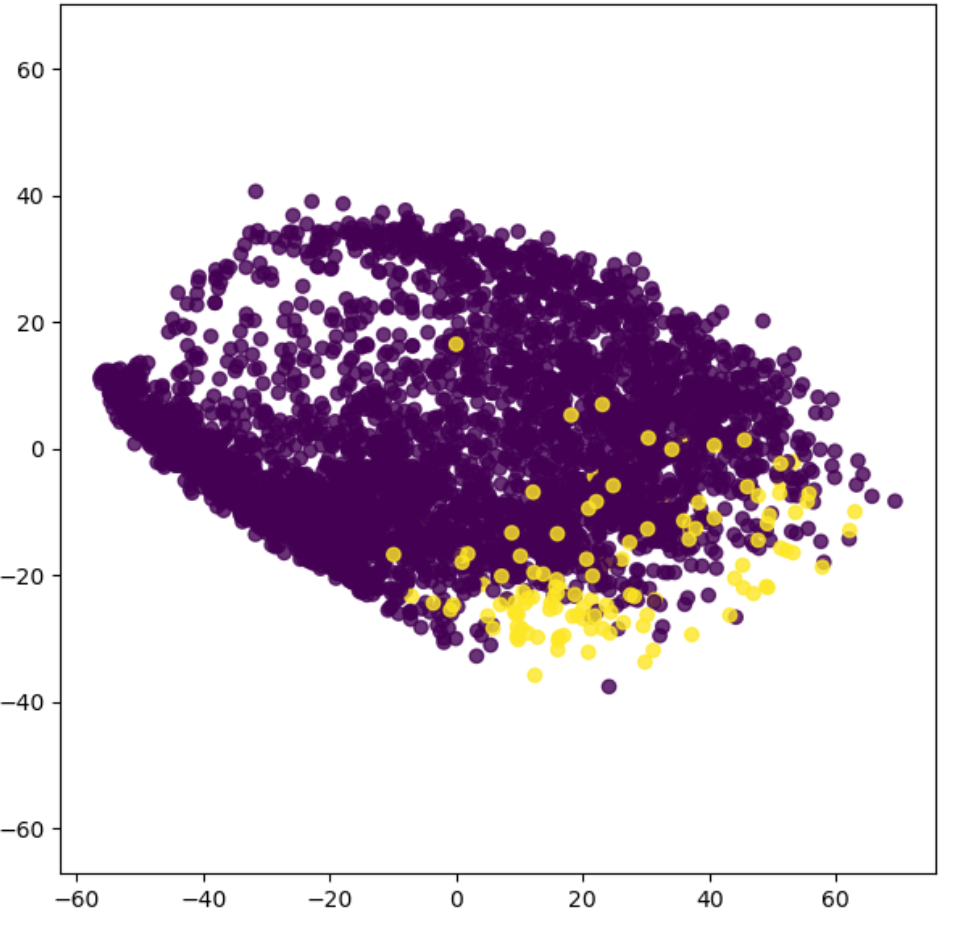}
        \caption{PCA}
        \label{fig:pca}
    \end{subfigure}
    \caption{Barlow Twins embedding space of LiDAR-CAT3 test data using dimensionality reduction techniques for visualization. Purple indicates point clouds exclusively composed of vegetation, while yellow denotes point clouds containing minority classes.}
    \label{fig:umap_pca}
\end{figure}

\begin{figure}[h]
\centering
\includegraphics[width=\columnwidth]{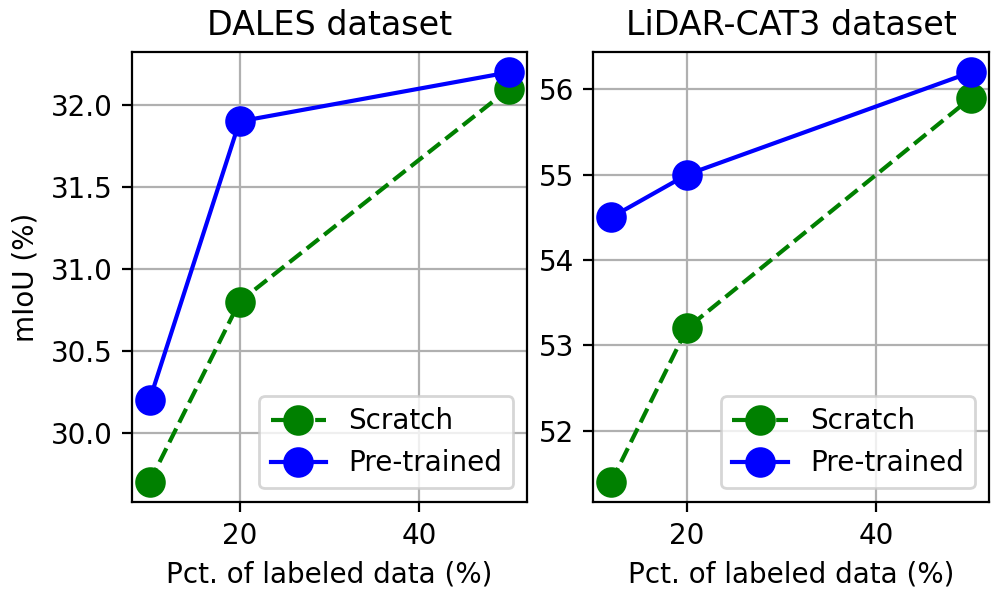}
  \caption{Result for different percentages of labeled data. Our pre-training method exhibits superior performance compared to training from scratch on both datasets.
  }
\label{fig:comparison_datasets}
\end{figure}

In this section, we perform several evaluations of the effectiveness of our method on DALES and LiDAR-CAT3  datasets for the task of 3D semantic segmentation and show that it learns meaningful representations. 
The results of the experiments are presented in Fig~\ref{fig:umap_pca}, Fig~\ref{fig:comparison_datasets}, and  Tables~\ref{tab:iou_categories} and  \ref{tab:dales_iou}. As evaluation metrics we use intersection-over-union per category (IoU), mean IoU (mIoU), and overall accuracy (OA).

In Fig~\ref{fig:comparison_datasets}, we show our experiment varying the fraction of labeled data up to 50\%. Remarkably, our pre-trained models outperform PointNet trained from scratch, especially when fine-tuning on fewer training samples. 

In Table~\ref{tab:iou_categories}, models were trained on 12\% of labeled data from our experimental dataset LiDAR-CAT3. We can see that pre-trained PointNet (SSL + PN) improved mIoU by 3.1 absolute points which is an increase of 6\% over the baseline. However, the highest mIoU score is achieved by the pre-trained PointNet\texttt{++} with a mIoU score of 61.2\%. Notably, our method performed particularly well in under-represented categories such as wires and roofs, increasing their IoU score by 26\% and 9\% respectively when using SSL+PN, and 7\% and 4\% when using PointNet\texttt{++}(SSL + PN\texttt{++}). To prove that our method generalizes to other datasets, we test our method on DALES. Table\ref{tab:dales_iou} presents the results for categories with IoU $>1\%$. We see that SSL pre-training in a low-data regime (10\% labeled data) improves performance in all categories.

\begin{table}[!h]
  \caption{Comparison between our proposed method and point-based architectures trained from scratch on LiDAR-CAT3 dataset using 12\% of labeled data.}
  \begin{center}
  \resizebox{\columnwidth}{!}{%
    \begin{tabular}{c | c c c c c c | c c}
      \hline
      \hline
      \makebox[10mm]{Methods} &
      \multicolumn{1}{p{6mm}}{\centering low \\ veg.} &
      \multicolumn{1}{p{6mm}}{\centering high \\ veg.} &
      \multicolumn{1}{p{6mm}}{\centering pylon} &
      \multicolumn{1}{p{6mm}}{\centering other \\ build.} & 
      \multicolumn{1}{p{6mm}}{\centering wires} & 
      \makebox[10mm]{roofs} &
      \multicolumn{1}{p{6mm}}{\centering mIoU \\ (\%)} & 
      \multicolumn{1}{p{6mm}}{\centering OA \\ (\%)} \\
      \hline
        PN         &  \textbf{72.2} & 82.2 & 46.8 & 13.3 & 31.3 & 62.7 & 51.4 & 91.5 \\
        SSL + PN   &  71.4 & \textbf{83.1} & 48.1 & 16.2 & 39.5 & 68.4 & 54.5 & \textbf{91.8} \\
      \hline
        PN\texttt{++}  &  62.4 & 82.2 & \textbf{66.2} & 20.4 & 56.8 & 72.9 & 60.2 & 90.7 \\
        SSL + PN\texttt{++} &  61.6 & 82.3 & 65.9 &     \textbf{20.7} & \textbf{61.1} & \textbf{75.6} & \textbf{61.2} & 90.8 \\
      \hline
      \hline
    \end{tabular}}
    \label{tab:iou_categories}
  \end{center}
\end{table}

\begin{table}[!h]
  \caption{Semantic segmentation results on Dales dataset using PointNet (PN) and pre-trained PointNet (SSL+PN) with our method using 10\% of labels.}
  \begin{center}
  \resizebox{\columnwidth}{!}{%
    \begin{tabular}{c | c c c c c | c c}
      \hline
      \hline
      \makebox[15mm]{Methods} &
      \makebox[7mm]{ground} &
      \makebox[5mm]{veg} &
      \makebox[5mm]{car} &
      \makebox[5mm]{wires} &
      \makebox[5mm]{build.} &
      \makebox[15mm]{mIoU (\%)} &
      \makebox[10mm]{OA (\%)} \\
      \hline
        PN       & 84.9 & 57.2 & 55.9 & 4.3 & 65.2 & 29.7 & 87.6 \\
        SSL + PN & \textbf{85.4} & \textbf{57.8} & \textbf{56.4} & \textbf{6.1} & \textbf{66.5} & \textbf{30.2} & \textbf{88.1} \\
      \hline
      \hline
    \end{tabular}}
    \label{tab:dales_iou}
  \end{center}
\end{table}

\section{Conclusions}

We propose an easy to implement SSL method for LiDAR data that works across point-based architectures. We show that models pre-trained with our method outperform those trained from scratch in a challenging task as it is aerial LiDAR semantic segmentation.

{\small
\bibliographystyle{unsrt}
\bibliography{paper}
}

\end{document}